# Comparing Large Language Model AI and Human-Generated Coaching Messages for Behavioral Weight Loss

**Zhuoran Huang**[1] · **Michael P. Berry**[2] · **Christina Chwyl**[3] · **Gary Hsieh**[4] · **Jing Wei**[5] · **Evan M. Forman**[6,7]

## Abstract

Automated coaching messages for weight control can save time and costs, but their repetitive, generic nature may limit their effectiveness compared to human coaching. Large language model (LLM) based artificial intelligence (AI) chatbots, like ChatGPT, could offer more personalized and novel messages to address repetition with their data-processing abilities. While LLM AI demonstrates promise to encourage healthier lifestyles, studies have yet to examine the feasibility and acceptability of LLM-based BWL coaching. Eighty-seven adults in a weight-loss trial (BMI ≥ 27 kg/m$^2$) rated ten coaching messages' helpfulness (five human-written, five ChatGPT-generated) using a 5-point Likert scale, providing additional open-ended feedback to justify their ratings. Participants also identified which messages they believed were AI-generated. The evaluation occurred in two phases: messages in Phase 1 were perceived as impersonal and negative, prompting revisions for messages in Phase 2. In Phase 1, AI-generated messages were rated less helpful than human-written ones, with 66% receiving a helpfulness rating of 3 or higher. However, in Phase 2, the AI messages matched the human-written ones regarding helpfulness, with 82% scoring three or above. Additionally, 50% were misidentified as human-written, suggesting AI's sophistication in mimicking human-generated content. A thematic analysis of open-ended feedback revealed that participants appreciated AI's empathy and personalized suggestions but found them more formulaic, less authentic, and too data-focused. This study reveals the preliminary feasibility and perceived helpfulness of LLM AIs, like ChatGPT, in crafting potentially effective weight control coaching messages. Our findings also underscore areas for future enhancement.

**Keywords** MHealth · Weight loss · Artificial intelligence · Large language model · ChatGPT

✉ Zhuoran Huang
huang.zhuor@northeastern.edu

[1] Khoury College of Computer Sciences, Northeastern University, 440 Huntington Ave, Boston, MA 02115, USA

[2] VA Connecticut Healthcare System, West Haven, CT 06516, USA

[3] Portland Psychotherapy Clinic, Research and Training Center, 3700 N Williams Ave., Portland, OR 97227, USA

[4] Department of Human Centered Design & Engineering, University of Washington, Seattle, WA 98195, USA

[5] School of Computing and Information Systems, The University of Melbourne, Parkville, VIC 3052, Australia

[6] Center for Weight, Eating, and Lifestyle Science, Drexel University, 3141 Chestnut Street, Stratton Hall, Philadelphia, PA 19104, USA

[7] Department of Psychological and Brain Sciences, Drexel University, 3141 Chestnut Street, Stratton Hall, Philadelphia, PA 19104, USA

## Introduction

Around 40% of adults worldwide (World Health Organization, 2021), and more than 70% in the USA (National Institute of Diabetes and Digestive and Kidney Diseases, n.d.), meet the criteria for overweight or obesity, posing risks such as type 2 diabetes, cardiovascular diseases, and cancer (Mokdad et al., 2003). Weight losses of 5% or greater can significantly mitigate these risks (Ryan & Yockey, 2017; Williamson et al., 2015).

Automated messaging, a technique within mobile health (mHealth; i.e., using phones and wearables for health interventions), is emerging as a scalable and efficient solution to address various health domains. Automated messages have been successfully incorporated into smoking cessation, diabetes management, and physical activity promotion programs, where they have been found to improve intervention engagement, increase self-management behaviors, and overall enhance clinical outcomes (Arora et al., 2014;

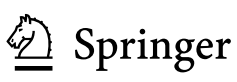

Buchholz et al., 2013; Scott-Sheldon et al., 2016). In the context of weight control interventions, automated coaching messages have shown promise for increasing efficacy, especially when they are integrated into comprehensive behavioral weight loss (BWL) programs (Anderson et al., 1999; Berrouiguet et al., 2016; Cavero-Redondo et al., 2020; Flores Mateo et al., 2015; Foster et al., 2005; Hernan et al., 2003; Kramer et al., 1989; Krukowski et al., 2011; Kuehn, 2022; Siopis et al., 2015; Skinner et al., 2020; Wadden et al., 1989; Wilson, 1994). However, the independent weight loss efficacy of automated messages, when not delivered as part of a comprehensive program, is typically quite low, i.e., 1–2% weight loss (Job et al., 2018; Skinner et al., 2020). A possible explanation for this finding is that the messages used in most prior interventions have been non-tailored, consisting of pre-drafted messages to provide weight loss tips, data summaries, or periodic reminders without individualization (Berry et al., 2023a; Partridge et al., 2020; Shaw & Bosworth, 2012). For example, a meta-analysis of text messaging-based intervention for health promotion found that tailored systems had larger effects than non-tailored ones (Head et al., 2013). Moreover, participants in weight loss interventions express a preference for personalized messages (Lyzwinski et al., 2018).

Tailored messages have the advantage of offering information that is more relevant to an individual user's needs. Still, a recent study by our research group reveals these messages often suffer from repetitiveness, impersonal tones, and redundant content, partly because they rely on a rule-based approach, where predefined rules lead to predictable content (Berry et al., 2023b). As a result, participants exhibit only a moderate level of satisfaction with such messages. Furthermore, the substantial costs and resources required to build such detailed systems from the ground up can impede their scalability.

Artificial intelligence (AI) systems, particularly large language models (LLMs), can understand and generate natural language through machine learning, transcending the constraints of rule-based systems (OpenAI, 2022). Users initiate conversations with LLMs through prompts, leading to model-generated responses. Using LLM AI can enhance personalization, reduce repetition, and foster increased novelty in content. For example, ChatGPT, an LLM chatbot from OpenAI, rapidly became popular after its 2022 launch. It hit 100 million active users in just two months, owing to its human-like interactions and vast knowledge base (Hu, 2023). A growing body of research has explored LLM AI's use cases in healthcare, such as engaging in medical writing and answering healthcare questions (Cascella et al., 2023; Sallam, 2023; Vaishya et al., 2023). More specifically, recent research has examined ChatGPT's performance in generating suggestions to optimize clinical decision support and answering questions about bariatric surgery (Liu et al., 2023; Samaan et al., 2023). Both studies demonstrate great potential for ChatGPT to serve as a helpful adjunct information resource for healthcare professionals.

Moreover, a developing body of research has expressed interest and advocated for further exploration of AI, including ChatGPT's capacity to deliver tailored obesity treatment and behavioral modifications (e.g., providing individualized advice on nutrition, exercise programs, and psychological support) (Arslan, 2023; Bays et al., 2023). Specifically, one study has indicated that GPT-3, an LLM AI and the precursor to ChatGPT, can aid in gathering self-reported data related to behaviors like food consumption and physical activity when using appropriate prompts (Wei et al., 2024). Such findings hint at the future role of AI chatbots in promoting healthy habits, from meal planning to fitness goal adherence. Given ChatGPT's ability to easily produce intricate messages with clinical nuance based on natural language inputs from patients or clinicians, it might generate personalized weight loss treatment messages without incurring extra tuning and development costs.

Despite the promising potential of LLMs like ChatGPT in diverse applications, there is a conspicuous absence of research examining the feasibility of LLM AI in crafting clinically relevant messages tailored for weight loss coaching. No studies, to date, have compared the perceived helpfulness of weight loss coaching messages generated by human experts and LLM AIs. This significant gap in research hinders the comprehension of LLM AI's potential to enhance the effectiveness of weight loss coaching or even substitute human coaches in specific clinical contexts.

In response to this research gap, our study explores the feasibility of leveraging ChatGPT to generate coaching messages and assesses the perceived helpfulness of the messages crafted. As an exploratory aim, we sought to understand the quality of the messages (e.g., language tone and fluency) and how the messaging content could be improved. We obtained qualitative feedback from weight loss-seeking participants who rated the helpfulness of messages produced by both AI and humans. Additionally, participants chose which messages they believed were written by ChatGPT and shared the strategies they used to identify them. We had two hypotheses: (1) that it would be feasible to generate messages using LLM AI and (2) that LLM AI messages would be rated as comparably helpful to human-generated messages.

To our knowledge, this is the first study to examine the feasibility of generating weight loss coaching messages using an LLM AI system and compare the perceived helpfulness between clinician-generated vs. AI-generated coaching messages in a clinical sample. The findings could carry considerable clinical implications for incorporating the LLM AI system into future personalized and cost-effective BWL strategies.

## Material and Methods

### Parent Trial

The data in the current study were collected from an ongoing behavioral weight-loss clinical trial, Project ReLearn (NCT05231824), which received approval from the Drexel University Institutional Review Board. In this year-long clinical trial, adults living in the USA who are overweight or obese receive weekly gold-standard behavioral weight loss interventions, which can be a small video conference group, a brief individual video call, or an automated coaching message. A previously published protocol paper reports additional details about the design, rationale, and eligibility criteria for the parent trial (Forman et al., 2023).

### Participants

Participants enrolled in Project ReLearn are adults aged between 18 and 70 living with overweight or obesity (BMI 27–50 kg/m$^2$). The current paper includes data obtained from 87 participants active in treatment during data collection. Participants were surveyed either in the 1-month (1 month from baseline, $N=47$) or the end-of-treatment assessment (12 months from baseline, $N=40$). Informed consent was obtained from all participants included in this study.

### Study Design

The study was carried out in two distinct phases: Phase 1 and Phase 2. In Phase 1, the same prompt was used to create human coach and ChatGPT messages. Participants reviewed ten messages, half from ChatGPT and half from humans, rating their helpfulness and substantiating their ratings with qualitative feedback. Moreover, to gauge the quality and natural tone of the AI to mimic human coaching, participants were prompted to discern which messages they believed the AI wrote, sharing the reasoning behind their choices. ChatGPT prompts were refined, and new AI messages were generated for Phase 2. In Phase 2, participants again rated helpfulness, discerned which messages were AI-written, and provided qualitative feedback on the original human messages and the revised ChatGPT messages. The study aimed solely to measure participants' subjective perceptions of helpfulness without attempting to correlate these perceptions with their weight loss outcomes.

### Message Generation

To generate coaching messages, we selected five scenarios (two weight loss, two weight gain, and one weight maintenance in the past week) from participants' data in the previous wave of the parent trial. Based on behavior change strategies supported in behavioral weight loss (Berry et al., 2023a; The Diabetes Prevention Program (DPP): description of lifestyle intervention, 2002), both human and ChatGPT messages summarized data patterns, praised aspects of the program that were progressing well, highlighted areas needing improvement, and offered constructive strategies to address those areas. This approach was applied to both weight change and one of four weight-related behavioral domains: physical activity, food tracking, self-weighing, or calorie management. Therefore, data including program week number, percent weight change since the program started, number of days weight tracked in the past week, participant's calorie goal range, and past three weeks' behavioral adherence data (weight change, days above/below/within calorie goal range, and physical activity goal and minutes) were provided to the human coach/ChatGPT for the chosen scenarios.

#### Human Message Generation Procedure

A highly-trained Master's-level weight loss coach (MB) with years of experience leading behavioral weight loss groups wrote a message for each scenario. The human coach had never read the messages created by the ChatGPT. The data from the five scenarios were presented in a table. The coach was provided with basic program information and the program's objective and was then instructed to write a five-sentence maximum message for each of the five scenarios based on the approach described above. See Fig. 1 for an example. All human-generated messages used in this study are listed in Supplementary Table S1.

#### AI Message Generation Procedure

We used ChatGPT (GPT-3.5) to generate the AI messages from the OpenAI web portal. A prompt structure based on the approach described above (similar to the one given to the human coach) generated ChatGPT's messages. Findings from the research show that specifying an identity (e.g., you are a weight loss coach) can improve the model performance of ChatGPT (Austin et al., 2021). Therefore, we crafted the ChatGPT prompt as follows: (1) informed the Chatbot that its role is to serve as a weight loss coach in a 52-week behavioral weight loss program, (2) provided the Chatbot with information that it needed to incorporate into a weight loss coaching message for a hypothetical participant, and (3) provided explicit instruction on the desired message format and content. An example of using ChatGPT to generate coaching messages is depicted in Fig. 2.

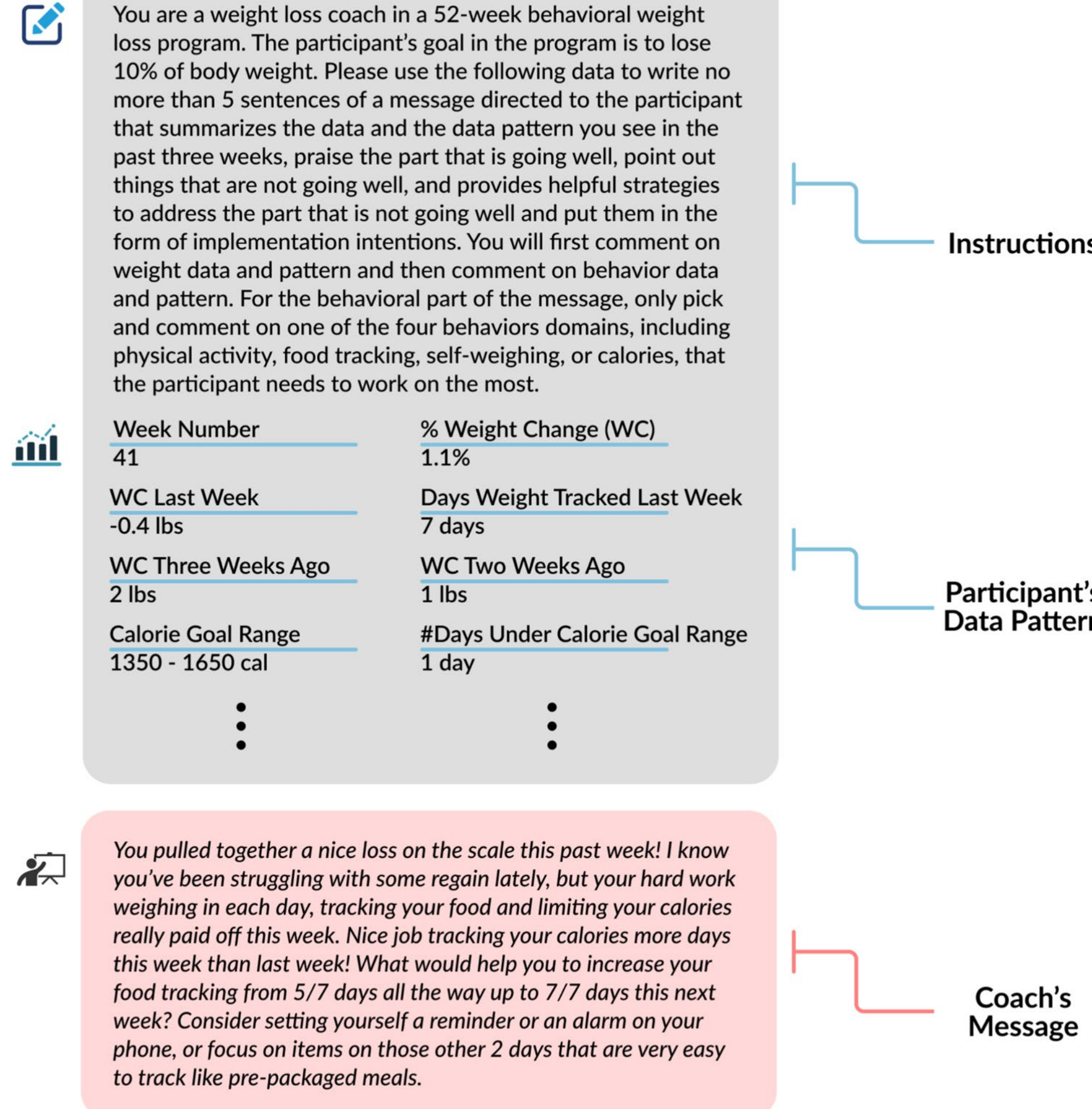

**Fig. 1** Example of how the human coach's message was generated

## Phase 1

In Phase 1, a similar instruction provided to humans was used to generate AI messages, as illustrated in Fig. 2. All prompts and messages from this phase can be found in Supplementary Table S2. A research coordinator in our team with prior qualitative data coding experience informally analyzed participants' feedback to discern their preferences and concerns regarding human and AI messages. This analysis revealed that ChatGPT messages, compared to human-written ones, (1) often sound more negative and impersonal with awkward phrasing, (2) tend to be overly data-driven, and (3) can sometimes make inaccurate assumptions.

## Phase 2

Given our Phase 1 findings, we adjusted the ChatGPT prompt in Phase 2 to mainly address the first concern noted above. We introduced tone modifiers in the Phase 2 prompt, such as "write a very encouraging and empathetic message with touches of humor," which were not specified in Phase 1 to align with the human coach's prompt. We also provided more detailed instructions, e.g., that the message should be in a single paragraph and use second-person pronouns. All Phase 2 prompts and messages can be found in Supplementary Table S3.

## Evaluation

### Acceptability Measures

To evaluate the perceived helpfulness of the messages written by the human coach and ChatGPT, all 87 enrolled participants completed a self-report survey and rated the helpfulness of ten messages. Participants were prompted to generate ideas for improving the effectiveness of the messages, blinded to the writer of each message and without being informed that some were written by AI. In the self-report survey, participants were presented with five scenarios, consistent with those coaches or ChatGPT, which were used to write messages. Within each scenario, participants were asked to look at a weekly data summary of the past week's weight, calorie, and activity information. They were then asked to imagine that these data summarized their weight control data patterns in the past week. Afterward, they saw two messages, one by the human coach and one by ChatGPT, with message order randomized within each scenario and were asked to rate the helpfulness of each message. Ratings were obtained on a 1–5 Likert scale where 5 represents a more favorable rating (1 = "Not at all helpful," 5 = "Extremely helpful"). Participants were also asked to provide free responses justifying their ratings.

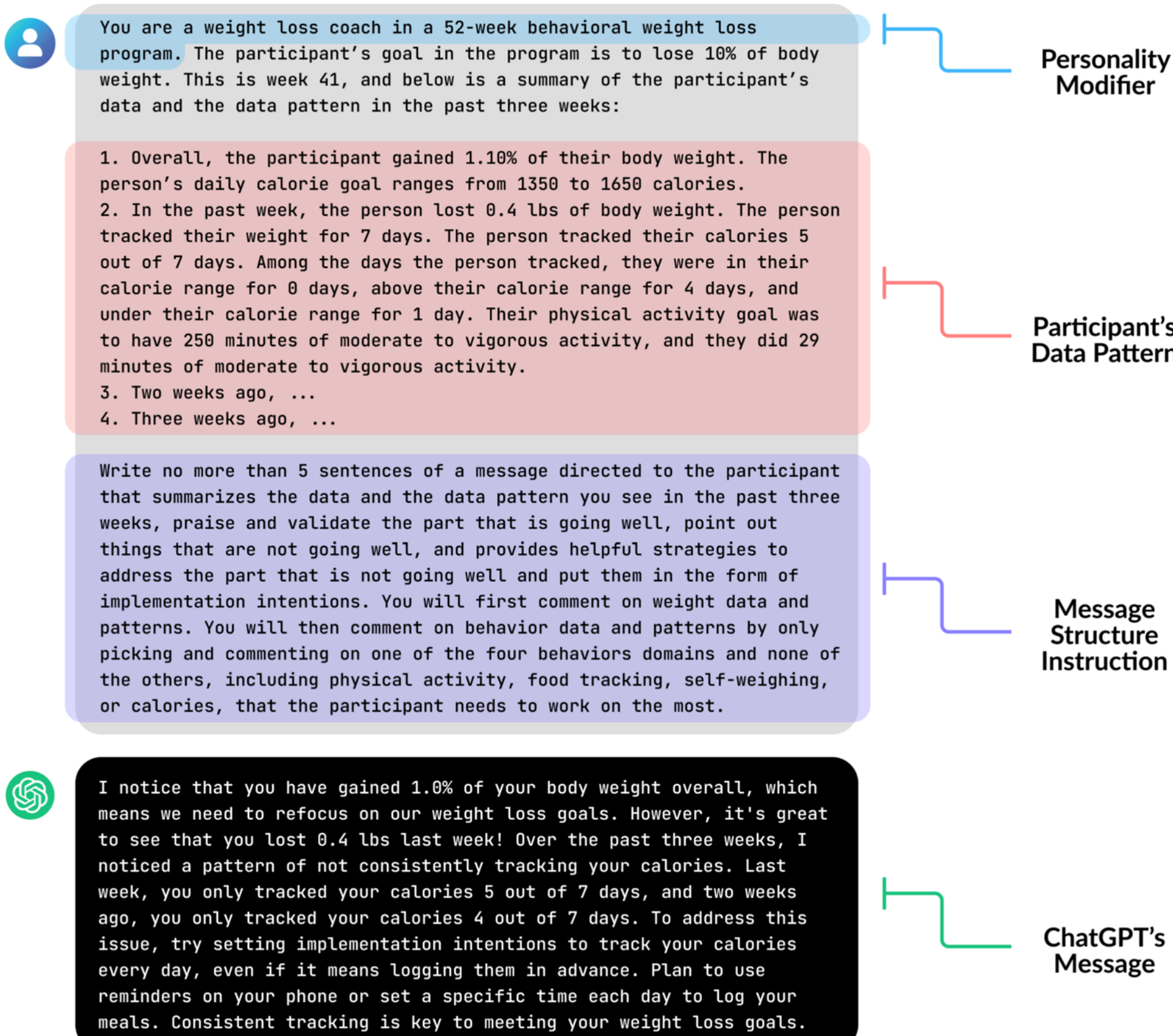

**Fig. 2** Example of ChatGPT prompt design and message generated

To evaluate the human-like quality of AI messages, we presented all ten messages in random order and prompted participants to identify which messages they believed were crafted by AI versus the human coach after they rated each of the ten messages. We also inquired about their differentiation strategies to discern the nuances in fluency and quality between human and AI communications.

### Data Analytic Strategy

We combined the ratings for each message category and calculated descriptive statistics of the helpfulness ratings. We applied the independent *t*-test to assess ChatGPT message improvement from Phase 1 to 2 and the paired-sample *t*-test to compare human coach and ChatGPT messages in both phases. We computed Cohen's *d* values to evaluate the effect size. We also looked at the accuracy of identifying ChatGPT-written messages.

Qualitative data collected via an electronic survey were analyzed using thematic analysis, a method for identifying and interpreting patterns across datasets (Braun & Clarke, 2006). Using an inductive approach, two authors (MB, ZH) derived themes directly from the data through a four-step process. They independently created succinct labels, or "codes," from significant data points, formulated overarching "themes" from these codes, collaboratively revised these themes, and refined them by revisiting the data, ensuring each code matched a theme. We only coded information related to participants' justifications for their rating, not the others. For free responses elucidating participants' strategies to discern message types, we evaluated only those with a correctness percentage at or above the median, focusing on participants who effectively differentiated between the two message types. Given the inductive and open-ended nature of the thematic analysis method we chose (Braun & Clarke, 2006), we aimed to ensure that our initial hypotheses did not constrain the resulting themes. This approach allowed us to capture and thoroughly describe the dataset, including participant responses that did not fully align with our hypothesis.

## Results

### Participant Characteristics

The participant sample in the study had an average age of 53.0 years (SD = 10.93, range 29–70) and an average baseline BMI of 34.32 (SD = 4.72, range 27.62–48.91) at enrollment. The self-reported racial distribution was as follows: 81.7% White, 8.0% Black, 4.6% Asian, and 4.6% multiracial; 1.1% did not report their race. Regarding ethnicity, 4.6% identified as Latino/Latina/Hispanic. The gender identity distribution was 81.6% female and 18.4% male. For employment, 60.9% of the participants were employed full-time, 24.2% were not working outside the home, such as being retired, 11.5% were employed part-time, and 3.4% received disability/SSI.

### Phase 1 Analysis

In Phase 1, ratings of helpfulness for human-written messages (median 4) were higher than AI-written ones (median 3, $t$ (234) = 8.41, $p < 0.001$), corresponding to a medium effect size ($d = 0.55$). Sixty-six percent of the AI-written and 89% of the human-written messages were rated as somewhat helpful to extremely helpful. 29.8% of the AI-written messages were misidentified as human-written (Table 1).

### Phase 2 Analysis

In Phase 2, ratings of helpfulness for human-written messages (median 4) were still significantly higher than AI-written ones (median = 4, $t$ (199) = 2.10, $p = 0.037$) but to a lesser degree, as evidenced by the small effect size of the difference ($d = 0.15$). Compared to Phase 1, AI-generated messages were rated as significantly more helpful ($t$ (433) = 4.97, $p < 0.001$), corresponding to a medium effect size ($d = 0.48$). However, human-written messages were not rated as more helpful in Phase 2 compared to Phase 1 ($t$ (433) = 0.54, $p = 0.59$). Eighty-two percent of the AI-written and 88% of human-written messages achieved an overall helpfulness score of 3 or higher (i.e., rated as somewhat helpful to extremely helpful). Fifty percent of AI-written messages were misidentified as human-written (Table 2). Notably, in the open-ended responses, many participants noted finding it challenging to differentiate between messages authored by humans and those generated by AI.

**Table 1** Accuracy in distinguishing AI from human-written messages in Phase 1

| | Messages written by human | Messages written by AI |
|---|---|---|
| Identified as human | 79.6% | 29.8% (incorrect) |
| Identified as AI | 20.4% (incorrect) | 70.2% |

**Table 2** Accuracy in distinguishing AI from human-written messages in Phase 2

| | Messages written by human | Messages written by AI |
|---|---|---|
| Identified as human | 62.5% | 50.0% (incorrect) |
| Identified as AI | 27.5% (incorrect) | 50.0% |

### Thematic Analysis Results

Qualitative feedback from both Phase 1 and Phase 2 was subjected to thematic analysis after the conclusion of the two phases. Because qualitative data were overall very similar across study phases, we presented data from both phases. The aim of the thematic analysis was not to shape the ChatGPT prompts but to understand participants' coaching message preferences, guiding potential future directions. Across the two study phases, we identified three themes: Theme 1 compares human-written vs. AI-generated messages, while the remaining themes pertain to both message types to offer insights on modifying prompts for future studies. Supplementary Table S4 presents a summary of themes with corresponding representative quotes.

#### Theme 1: AI-Generated Messages Feel More Formulaic and Impersonal, with Less Encouragement of Autonomy

Participants frequently commented that the AI-generated messages were more formulaic than those written by a human coach. For instance, a participant remarked: "They seemed more like filling in boxes and putting trite comments than a person with emotions. (1063)".

Furthermore, participants noted that AI-crafted messages often convey an inauthentic tone with awkward wording, characterized by excessive exclamation marks, enthusiasm, or a lack of first-person pronouns.

Contrarily, human-coach-written messages were perceived as more personal and empathetic. They were described as more personal and displaying a profound

"understanding of the struggles" and an ability to "recognize and encourage positive behaviors. (1029)".

A recurring observation was the officious nature of AI-generated messages, which are heavily based on user data trends rather than individual nuances. Consequently, participants sometimes perceive the AI's tone as overtly assertive or "bossy." One participant said: "[The message] has good motivation, almost too much that to me tilts to and almost patronizing. (1067)".

In contrast, human messages were more open-ended and collaborative, and one participant stated: "[The human coach's] message sounds curious ('It might be this…it could be related to this…'), which, to me, feels engaging and respectful of my competence and self-awareness. (1008)".

Such client-focused messages, which encourage individuals to reflect on the feedback and develop their own personalized action plans, were favored by participants across message types.

Figure 3 illustrates that the human coach's message promotes greater autonomy by prompting participants to identify methods for calorie reduction and offering tangible examples, whereas ChatGPT's message is very instructional.

### Theme 2: Participants Desire Messages that Offer Validation and Motivational Support, Complemented by Specific and Personalized Recommendations for Behavioral Change

Participants consistently emphasized the importance of encouragement and affirmation within coaching messages, particularly during challenging times. One participant shared:

> I have had weeks like this where I am feeling pretty bummed about my progress, and getting a message like this is so encouraging and motivating for me to want to continue the program even if I am not seeing the loss that I would like to see. (1023)

Yet, while validation was desired, an overemphasis on encouragement led some participants to feel that messages resembled "cheerleading" and were overly enthusiastic. Instead of unchecked positivity, there was a noted preference for messages to strike a balance or even offer more "push"—affirming achievements while candidly highlighting areas of growth.

Besides the need for a balanced tone and validation, many participants underscored the need for coaching messages to be more concise. While brevity was appreciated, it should not be prioritized at the expense of personalization or actionable insights. Messages that seemed repetitive or echoed readily available information were less valued than those that offered clear and specific guidance. One participant said:

> [The message] is positive and sets an intention, but it needs to be more specific. i.e., "I will track daily and aim to consume 'X' calories daily. I will do that by cutting out sugar in my coffee and eating while watching Netflix." (1040)

Meanwhile, participants consistently desired messages that effectively synthesize, distill, or condense their data, making it understandable and actionable. Furthermore, they frequently expressed a need for more comprehensive explanations within the coaching messages regarding the data's implications and a desire for the data to be referenced more consistently in supporting the advice given. One participant shared: "[The message] need[s] to look at the calorie data being tracked and do some analysis—is there a pattern that can be discerned that perhaps the participant is unaware of? (3027)".

**Fig. 3** Example of human coach and ChatGPT's messages

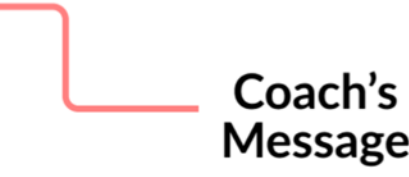

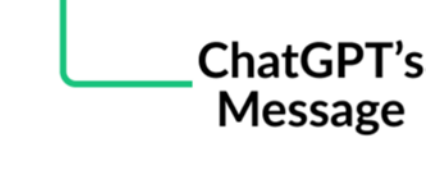

Lastly, many participants noted that the messages served as valuable reminders of their behavioral goals and the skills they acquired during group treatment sessions, thereby aiding them in maintaining their progress.

### Theme 3: Participants Seek Messages to Consider the Full Context of Their Data Trends to Offer More Targeted Behavior Suggestions

Several participants expressed that the comments on their data should also consider the broader context of their weight loss journey without making assumptions. For example, participants suggested that messages should explore other potential reasons besides the shortcomings in one domain, like unrealistic goals or difficulty tracking calorie intake during travel, to offer more targeted strategy suggestions. A participant, rating a message as 3 out of 5, emphasized the issue with assumptive components and contextual inadequacy by noting, "I think (for me) the problem is the assumption component and lack of context. What if this person were on vacation—that would be a great week. (1014)".

Additionally, another participant complained that the message (see Fig. 4) lacked consideration of the contextual factors explaining their inability to track calories daily, rendering the advice provided less than optimal: "[The message] would be a good reminder of how important tracking is but may not get at the reason why tracking wasn't completed. For example, an illness, family emergency, etc. (3020)."

Moreover, another participant highlighted that the message (see Fig. 5) did not take situational factors such as the time of the program (i.e., week 39 out of 52) into account and, therefore, providing feedback that is general and lacked salience in promoting positive behavioral change:

> At Week 39, with an overall gain, I'd suspect that unless I had something medically or emotionally going on with me causing this gain, I'd simply lose interest in the program. Just telling me to track at this point without supportive suggestions wouldn't be motivating. I would probably not change my habits as a result of this message if I hadn't already. (3062)

**Fig. 4** Example message 1 for Theme 3

I notice that you have gained 1.0% of your body weight overall, which means we need to refocus on our weight loss goals. However, it's great to see that you lost 0.4 lbs last week! Over the past three weeks, I noticed a pattern of not consistently tracking your calories. Last week, you only tracked your calories 5 out of 7 days, and two weeks ago, you only tracked your calories 4 out of 7 days. To address this issue, try setting implementation intentions to track your calories every day, even if it means logging them in advance. Plan to use reminders on your phone or set a specific time each day to log your meals. Consistent tracking is key to meeting your weight loss goals.

**Fig. 5** Example message 2 for Theme 3

Great job on tracking your weight for the past three weeks! Unfortunately, the data shows that you have gained weight in the past week and have had an overall gain of 0.7% of your body weight. To improve your weight loss journey, it's important to focus on tracking your food intake accurately. Implement the strategy of tracking your calories every day, aiming to be within your calorie range of 1350 to 1650 calories. This will help you to stay on track and reach your goal of losing 10% of your body weight.

## Discussion

Automated coaching messaging offers a cost-effective mHealth solution for weight loss, but many systems lack personalization, potentially limiting outcomes. LLM AI has the potential to craft tailored coaching messages inexpensively, but its feasibility and effectiveness remain unexplored in weight-loss contexts. In this study, 87 participants seeking weight loss assessed the helpfulness of ten coaching messages—five from a human and five from ChatGPT—using a 5-point Likert scale and provided feedback. They also identified messages they believed were AI-generated. We proposed that (1) creating messages with LLM AI could be feasible and (2) messages generated by LLM AI would receive ratings of helpfulness similar to those of human-generated messages. We aimed to assess the message quality, including its tone and fluency, and explore ways to enhance the content as an exploratory aim. This evaluation occurred in two phases, with the second phase building on the first's feedback.

In the first phase, ChatGPT successfully generated weight loss coaching messages by interpreting step-by-step instructions in plain English and summarizing data-rich content. However, it initially struggled with capturing the nuanced tone required for effective coaching. With a refined prompt in Phase 2, the messages displayed support and empathy and offered personalized behavioral suggestions. We showed that ChatGPT-generated messages were just as helpful as human-authored ones and received a helpfulness rating of 3 or more in 82% of cases, a significant increase compared to the 60% rating observed for the rule-based messaging system used in the parent study (Berry et al., 2023b). Of note, participants were no better than chance at identifying the authorship of AI messages. The number suggests that AI messages can effectively convey natural empathy while providing specific and inspiring suggestions, though further refinement remains necessary. Following a marked increase in helpfulness ratings due to a simple prompt redesign, our study emphasizes ChatGPT's adaptability and potential for crafting increasingly personalized messages. As demonstrated in our study, the rapid advancement of LLM AI has promising implications for enhancing the cost-effectiveness of hybrid or fully automated BWL programs. This could significantly save clinicians time and financial resources for tailored messaging. For example, GPT-4, OpenAI's latest LLM, has made notable improvements in solving complex tasks like synchronizing overlapping schedules (OpenAI, 2023). This model is more reliable, creative, and capable of handling nuanced instructions, offering greater steerability for users to prescribe both a specific task and a personality. Hence, future LLM AI models should excel at following pre-defined roles and instructions, consistently showing empathy as a weight loss coach, discerning clinical nuances, and offering precise feedback.

Given the promise of GPT-4, we believe it can address some concerns highlighted in our thematic analysis. For instance, messages from ChatGPT can sometimes feel formulaic, less authentic, and overly prescriptive compared to those from humans. Our findings, consistent with prior research, show participant preference for positive, supportive, personalized, and jargon-free messages (Lyzwinski et al., 2018). In post hoc tests following the main study, GPT-4, with optimized prompts, produced more balanced and collaborative messages, incorporating Socratic questioning and respecting individual autonomy. This underscores the potential to narrow the disparity between human-written messages and those produced by LLM AIs like GPT-4.

Themes applicable to both message types highlighted the strengths of employing LLM AI in message creation. Compared to traditional rule-based systems, LLM AI's flexibility opens the door for more customized messaging. Recognizing that participants have varied tone preferences, future systems could introduce a tone slider, enabling users to select between more assertive or encouraging messages and choose areas for specialized feedback. Future LLM AI systems could allow participants to share more information about the situational context (e.g., sick or traveling) to have it offer more individual personalized suggestions.

The current study underscores the potential clinical advantages of incorporating an advanced AI system into mobile BWL, particularly given its demonstrable feasibility and high perceived helpfulness. This integration could also be instrumental in conserving clinical resources in the fight against the worldwide obesity crisis. Future research should focus on refining prompts to continually optimize the content and further evaluate the clinical effectiveness of AI-generated messages in randomized controlled trials.

However, incorporating LLMs into research and clinical practice raises critical ethical concerns, particularly regarding data privacy and compliance with regulations like the Health Insurance Portability and Accountability Act (HIPAA). LLMs often require substantial data inputs, including sensitive health information, to generate personalized responses, heightening the risk of exposing such data. Additionally, if not carefully monitored, AI systems may inadvertently generate inaccurate or misleading health advice or fail to respond to specific inquiries—challenges often associated with LLM AI (Bang et al., 2023; OpenAI, 2023). To mitigate these concerns, it is critical to implement robust safeguards, such as encrypting user data, restricting AI access to de-identified or minimal datasets, and conducting regular audits and features enabling users to seek human intervention to ensure the system is safe and adheres to privacy standards. Further, integrating user consent protocols and providing transparency about how data is processed can help build trust and accountability. Developing AI-specific training for clinicians and researchers could also enhance oversight, ensuring LLMs are applied responsibly in clinical settings.

## Limitations

The current study exhibits several limitations. First, the sensitivity of ChatGPT to provided prompts can lead to inconsistencies in generated messages. Although we devised the prompt format by examining various input structures, ChatGPT occasionally misinterpreted or inadequately addressed them, causing variations in message quality. Thus, some messages we used were not from ChatGPT's initial response but were picked from several outputs stemming from an identical prompt. Secondly, during phase 2 of the study, we focused on addressing the lack of encouragement and empathy in messages, neglecting other potential concerns. Therefore, increasing prompt specificity, such as introducing reflective questions or avoiding assumptions, may enhance the message's performance.

Additional limitations include that the messages received by participants were based on data from a hypothetical participant rather than their own, potentially compromising the study's ecological validity. Finally, participants self-selected for the parent study, an AI-based BWL, suggesting they might be more open to AI-generated messages, potentially limiting the study's generalizability.

# Conclusion

Overweight and obesity are global concerns. Our study evaluated ChatGPT's ability to generate weight loss coaching messages and compared it to a human coach. The AI-generated messages exhibited moderate to high acceptability for their helpfulness and appeared to match human messages in content and language closely. Thematic analysis showed that while AI messages conveyed empathy and encouragement and gave targeted weight management advice, they often felt formulaic and overly data-driven. Our study highlights LLM AI's potential to enhance future weight loss interventions, making them more personalized, scalable, and cost-effective. We expect these insights to drive further research on LLM AI methods, potentially addressing the global obesity crisis.

**Supplementary Information** The online version contains supplementary material available at https://doi.org/10.1007/s41347-025-00491-5.

**Acknowledgements** We extend our gratitude to Nikoo Karbassi, a research coordinator at our center, for her qualitative data analysis during the first study phase. We also thank Charlotte Hagerman, Ph.D., for her expert consultation on the data analysis plan.

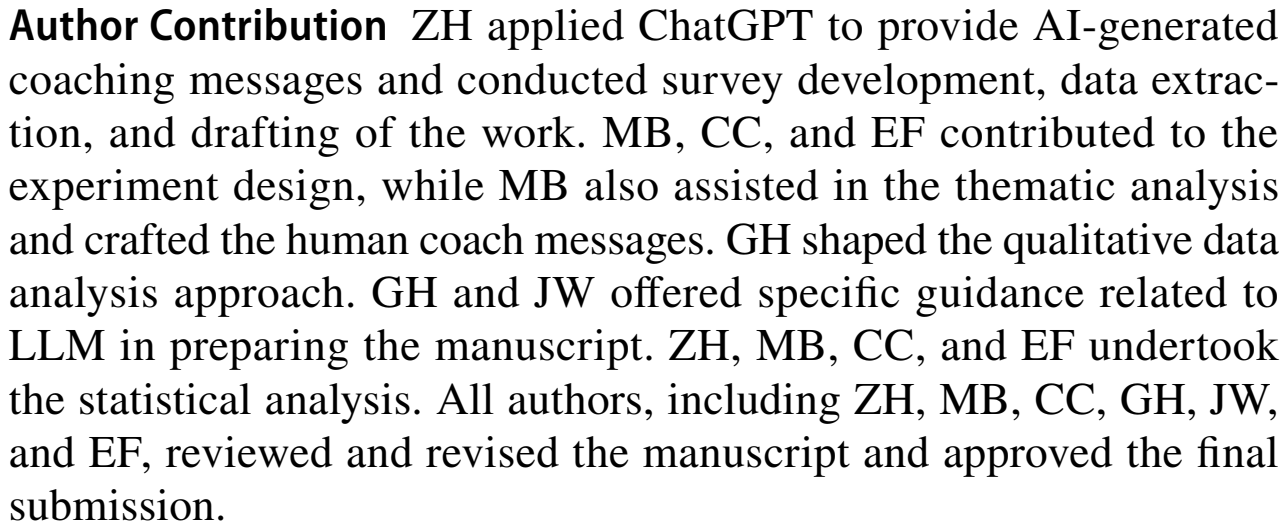

**Author Contribution** ZH applied ChatGPT to provide AI-generated coaching messages and conducted survey development, data extraction, and drafting of the work. MB, CC, and EF contributed to the experiment design, while MB also assisted in the thematic analysis and crafted the human coach messages. GH shaped the qualitative data analysis approach. GH and JW offered specific guidance related to LLM in preparing the manuscript. ZH, MB, CC, and EF undertook the statistical analysis. All authors, including ZH, MB, CC, GH, JW, and EF, reviewed and revised the manuscript and approved the final submission.

**Funding** Open access funding provided by Northeastern University Library. This work was supported by NIH grant number R01DK125641.

**Data Availability** The input prompts, AI outputs, themes, and representative extracts are available in the Supplementary Appendix. Other data that support the findings of this study are available from the authors upon reasonable request and with the permission of the study center.

## Declarations

**Conflict of Interest** The authors declare no competing interests.

**Disclaimer** All opinions and views expressed in this article are those of the authors and not those of the institution.